%% file: lrec-coling2024-example.tex
\documentclass[10pt, a4paper]{article}
\usepackage{lrec-coling2024}  

\usepackage{natbib}
\usepackage{multibib}
\makeatletter
\def\@mb@citenamelist{cite,citep,citet,citealp,citealt,citepalias,citetalias}
\makeatother
\newcites{languageresource}{~}

\usepackage{graphicx}
\usepackage{tabularx}
\usepackage{soul}
\usepackage{subfig}
\usepackage{subcaption}

\newcommand{\SmallTitleFont}{\normalfont\bfseries\raggedright}

\usepackage{titlesec}
\titleformat{\section}{\normalfont\large\bfseries\center}{\thesection.}{1em}{}
\titleformat{\subsection}{\normalfont\SmallTitleFont\bfseries\raggedright}{\thesubsection.}{1em}{}
\titleformat{\subsubsection}{\normalfont\normalsize\bfseries\raggedright}{\thesubsubsection.}{1em}{}
\renewcommand\thesection{\arabic{section}}
\renewcommand\thesubsection{\thesection.\arabic{subsection}}
\renewcommand\thesubsubsection{\thesubsection.\arabic{subsubsection}}

\usepackage{amsmath} 
\usepackage{amsfonts}
\usepackage{booktabs}
\usepackage{arydshln}
\usepackage[inline]{enumitem}
\usepackage{multirow}
\usepackage{threeparttable} 
\usepackage{xcolor}
\usepackage{hyperref}
 \definecolor{darkblue}{rgb}{0, 0, 0.5}
  \hypersetup{colorlinks=true, citecolor=darkblue, linkcolor=darkblue, urlcolor=darkblue}

\usepackage{xstring}

\usepackage{color}

\title{ 
Multimodal Cross-Document Event Coreference Resolution Using Linear Semantic Transfer and Mixed-Modality Ensembles
}

\name{Abhijnan Nath\textsuperscript{1}, Huma Jamil\textsuperscript{1}, Shafiuddin Rehan Ahmed\textsuperscript{2},
 George Baker\textsuperscript{2}, \\ 
 Rahul Ghosh\textsuperscript{1,3}$^{*}$, James H. Martin\textsuperscript{2}, Nathaniel Blanchard\textsuperscript{1}, and Nikhil Krishnaswamy\textsuperscript{1}}
 
\address{\textsuperscript{1}Colorado State University, Fort Collins, CO, USA\\
\textsuperscript{2}University of Colorado, Boulder, CO, USA\\
\textsuperscript{3}Purdue University, West Lafayette, IN, USA\\
\{abhijnan.nath, nkrishna\}@colostate.edu}

\abstract{
Event coreference resolution (ECR) is the task of determining whether distinct mentions of events within a multi-document corpus are actually linked to the same underlying occurrence. Images of the events can help facilitate resolution when language is ambiguous. Here, we propose a multimodal cross-document event coreference resolution method that integrates visual and textual cues with a simple linear map between vision and language models. As existing ECR benchmark datasets rarely provide images for all event mentions, we augment the popular ECB+ dataset with event-centric images scraped from the internet and generated using image diffusion models. We establish three methods that incorporate images and text for coreference: 1) a standard fused model with finetuning, 2) a novel linear mapping method without finetuning and 3) an ensembling approach based on splitting mention pairs by semantic and discourse-level difficulty. We evaluate on 2 datasets: the augmented ECB+, and AIDA Phase 1. Our ensemble systems using cross-modal linear mapping establish an upper limit (91.9 CoNLL F1) on ECB+ ECR performance given the preprocessing assumptions used, and establish a novel baseline on AIDA Phase 1. Our results demonstrate the utility of multimodal information in ECR for certain challenging coreference problems, and highlight a need for more multimodal resources in the coreference resolution space.
 \\ \newline  }

\begin{document}

\maketitleabstract

\def\thefootnote{*}\footnotetext{This work conducted at Colorado State University.}\def\thefootnote{\arabic{footnote}}

\section{Introduction}
\label{sec:intro}

Imagine two newspaper articles about the same event. The articles come from different sources with radically different perspectives and report the event with very different language. They use different action verbs, include ambiguous pronominal references, describe causes differently, and even attribute different intentionality to the event---for example, ``{\it Buzina, 45, \underline{was shot} dead}'' vs. ``{\it He was \underline{murdered}}''. An automated system may be unable to identify from the text alone that the two events described are actually the same.  This is the problem of {\it cross-document coreference resolution} (CDCR) of events: inferring that two event mentions in different documents actually refer to the same thing.

Now imagine that each of the articles is accompanied by an image. While not identical, they clearly contain the same people, entities, and actions. This would be strong evidence to a reader that the two events described in the different articles are in fact the same.

Purely text-based approaches to CDCR, while built on sophisticated Transformer-based language models (LMs)~\cite{vaswani2017attention,beltagy2020longformer}, are blind to such potentially useful multimodal information. This problem is exacerbated by the relative dearth of multimodal information included in event CDCR corpora.

In this work, we propose a novel multimodal {\it event} CDCR method. Where current state-of-the-art coreference approaches that consider visual information demonstrate the utility of a multimodal approach, they do so at a high computational cost~\cite{guo2022gravl}. Furthermore, they typically focus on linking objects rather than events. We address the sparsity of multimodal data in benchmark datasets by retrieving images associated with the metadata of event mentions, and generating event-centric images with state-of-the-art image diffusion models. We perform coreference experiments in a fully multimodal setting and rigorously test the contribution of multimodal information to CDCR.\footnote{Our code can be accessed at \url{https://github.com/csu-signal/multimodal-coreference}.}

In total, our novel contributions include:

\begin{itemize}
    \vspace*{-2mm}
    \item A novel approach to multimodal {\it cross document event coreference} (MM-CDCR) including a low-compute, bidirectional {\it linear semantic transfer} technique ({\tt Lin-Sem}) based on \textit{semantic equivalence} across modalities;
    \vspace*{-2mm}
    \item A model ensemble hybrid approach that applies text-only or multimodal methods to different categories of mention pairs based on their semantic and discourse-level difficulty;
    \vspace*{-2mm}
    \item A novel method for enriching text-only coreference datasets (e.g., ECB+~\cite{cybulska2014using}) with event-centric images using generative image diffusion;
    \vspace*{-2mm}
    \item A new benchmark result on the AIDA Phase 1 dataset \cite{tracey2022study}, an explicitly multimodal event CDCR dataset. To our knowledge, this is the first evaluation performed over this dataset.
    \vspace*{-2mm}
\end{itemize}


 \vspace*{-2mm}
\section{Related Work}
\label{sec:rel_work}
 \vspace*{-2mm}

\paragraph{Cross-Document Event Coreference Resolution}

Most previous works on CDCR have been limited to \textit{text-only} \cite{eisenstein-davis-2006-gesture, chen2011improving}. Early works (e.g.,~\citet{humphreys1997event,bagga1999cross,chen-ji-2009-graph}) used supervised training over features like part-of-speech tags, phrasal-matching, or aligned arguments. While \citet{kenyon2018resolving} enhanced lexical features with ``static'' embeddings like contextual word2vec \cite{mikolov2013efficient}, most recent works \cite{yu2022pairwise, caciularu-etal-2021-cdlm-cross, yadav2021event,nath-etal-2023-axomiyaberta} uses latent representations from Transformer-based encoders to compute pairwise mention scores of possible antecedents. Works such as \citet{held-etal-2021-focus} and \citet{ahmed20232} overcome the quadratic complexity of the mention pair architecture by pruning negative pairs using discourse-coherence and lexical similarity (synonymous lemma pairs) respectively. We use \citet{ahmed20232}'s ``oracle'' assumption for our pruning procedure. 





\paragraph{Multimodal Frameworks}

Most previous works in multimodal vision-language processing (e.g.,~\cite{le2019multimodal,tan2019lxmert}) have been compute-intensive, using separate encoders for visual and linguistic inputs, and auxiliary encoders for cross-modal or query-related modeling. High-performing but high-compute models like ViL-BERT~\cite{lu2019vilbert} concatenate embeddings from different modalities before fine-tuning.
Works such as \citet{li2020cross}, \citet{tong2020image}, and  \citet{chen2021joint} leverage a common representation space for coreference-adjacent tasks like event extraction and detection in images and videos, but emphasize finding relations within a document or a topic. Works specific to multi-modal {\it entity} coreference resolution such as \citet{guo2022gravl} treat it largely as a grounding problem, using graph networks to link references in dialogue to items in a scene before feeding representations into BERT-style encoders to resolve scene-based visual-linguistic coreference chains. Our work is multimodal, cross-document, {\it and} event focused, and performs faster with the aid of linear mappings. 

\paragraph{Linear Projection Across Neural Networks}

Previous research within computer vision has explored using affine~\cite{mcneely2020inception,mcneely2022canonical,10.3389/fcomp.2023.1274832} as well as non-linear~\cite{lenc2015understanding} transformations to explore equivalence of \textit{unimodal} function approximators like CNNs. They show that two distinct, highly non-linear neural networks can learn similar properties transferable up to a linear projection while retaining near-equivalent performance on tasks like image classification or facial recognition. Similar techniques using affine mappings were reported in \citet{merullo2023linearly}, who explore the equivalence of such approximators \textit{across} modalities while also casting new light on high-fidelity transfer of non-linguistic features into a generative LLM via unidirectional linear projections from image spaces. \citet{nath2022phonetic} demonstrated linear mappings also preserve information across language models. \citet{ghaffari2023grounding} showed the same between language models and neural networks trained over tabular data.


We use a low-compute, cross-modal, \textit{bidirectional} linear-mapping technique ({\tt Lin-Sem}: \textbf{Lin}ear \textbf{Sem}antic Transfer) between language and vision Transformers, on the challenging event coreference task. We demonstrate where this linear transfer is providing useful information toward coreference resolution compared to a text-only discriminative LLM, or fused modality models following standard fine-tuning. 

\begin{figure*}[t]
  \centering
  \includegraphics[width=\textwidth]{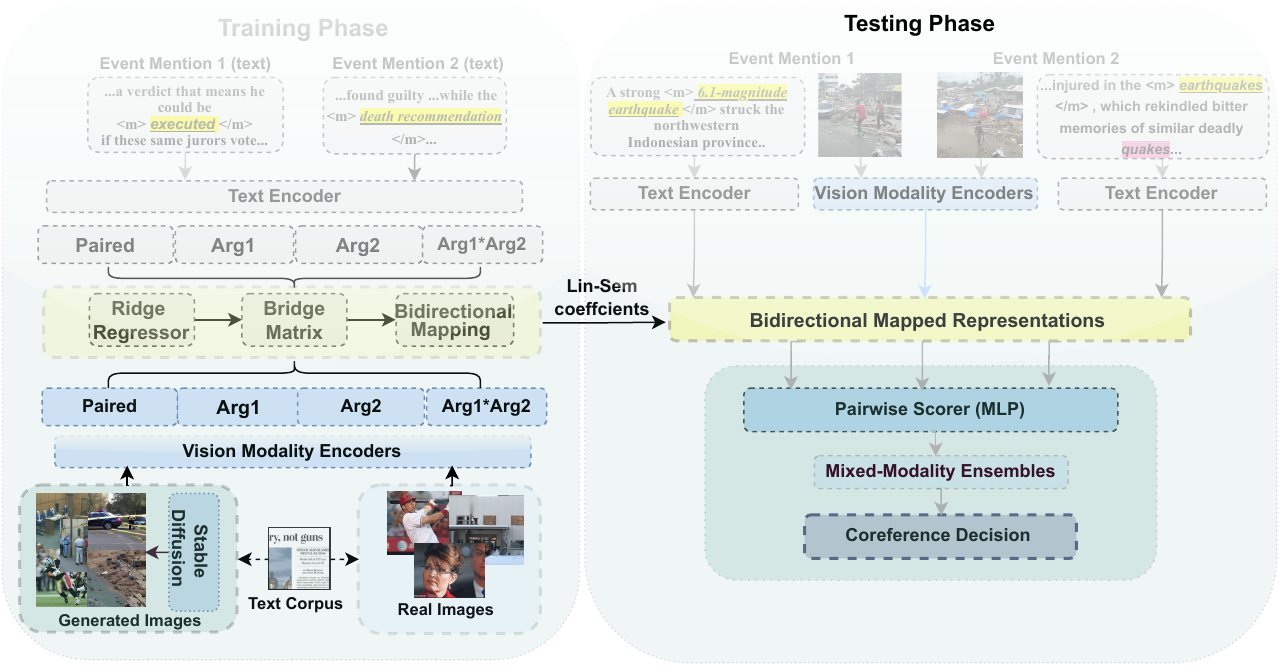}
 \vspace*{-2mm}
\caption{Our approach for Multimodal CDCR using {\tt Lin-Sem}. Linear Mapping ({\tt Lin-Sem}) procedure between the distinct text and image embedding spaces for an event pair in the ECB+ corpus. {\tt Arg1} and {\tt Arg2} refer to the individual images in the pair and the trigger events (in yellow) surrounded by the {\tt <m>} and {\tt </m>} special tokens embedded in the text-encoder (LLM).}  \label{fig:pipeline}
 \vspace*{-2mm}
\end{figure*}

 \vspace*{-2mm}
\section{Methodology}
\label{sec:method}
 \vspace*{-2mm}
\newcommand{\ViT}{\ensuremath{\texttt{ViT}}}
\newcommand{\BEiT}{\ensuremath{\texttt{BEiT}}}
\newcommand{\SWIN}{\ensuremath{\texttt{SWIN}}}
\newcommand{\CLIP}{\ensuremath{\texttt{CLIP}}}
\newcommand{\Random}{\ensuremath{\texttt{Random}}}
\newcommand{\LLMFT}{\ensuremath{\texttt{LLM-Finetuned}}}
\newcommand{\LLM}{\ensuremath{\texttt{LLM}}}

Fig.~\ref{fig:pipeline} illustrates the pipeline for our methodology, the components of which are detailed as follows.

\paragraph{Semantic Equivalence} 
 
\vspace*{-2mm}
\begin{equation} 
{\fontsize{8.6}{32}\selectfont
  \scriptstyle
  \mathcal{V}\bigl(x, y,\phi(x,y)\bigr):\mathbb{R}^{n\times w\times h\times 3}\rightarrow{} \mathbb{R}^{n\times H}
}
\label{eq:1}
\end{equation}
\vspace*{-4mm}
\begin{equation}
{\fontsize{8.6}{32}\selectfont
  \scriptstyle
  \mathcal{LLM}\bigl(x, y,\phi(x,y)\bigr):\mathbb{R}^{n\times m}\rightarrow{} \mathbb{R}^{n\times H}
}
\label{eq:2}
\end{equation}

Let (\ref{eq:1}) and (\ref{eq:2}) represent the heterogeneous image and text representations for vision and text Transformer models respectively. $(x,y) \in \chi$ represents all the pairs of samples in sample space $\chi$, $\phi(x,y)$ represents the concatenation of the image or text pair in their respective modalities, $n$ and $H$ represent the total sample pairs and hidden dimensions respectively, and $m$ is the LLM's max token-length.
 

We define cross-modal semantic equivalence as follows: two representations $\mathcal{V}$ and $\mathcal{LLM}$ in distinct modalities are semantically equivalent if there exists a bidirectional map $M_{\mathcal{V}\leftrightarrow{}\mathcal{LLM}}$ s.t.:
\vspace*{-2mm}
\begin{equation} 
{\fontsize{8.6}{32}\selectfont
  \scriptstyle
  \forall x,y \in \chi: \mathcal{V}\bigl(x, y,\phi(x,y)\bigr) \approx M_{\mathcal{LLM}\rightarrow{}\mathcal{V}} \mathcal{LLM}\bigl(x, y,\phi(x,y)\bigr)
}
\end{equation}
\vspace*{-4mm}
\begin{equation}
{\fontsize{8.5}{32}\selectfont
  \scriptstyle
  \forall x,y \in \chi: \mathcal{LLM}\bigl(x, y,\phi(x,y)\bigr) \approx M_{\mathcal{V}\rightarrow{}\mathcal{LLM}} V\bigl(x, y,\phi(x,y)\bigr)
}
\end{equation}
while assuming both $\mathcal{V}$ and  $\mathcal{LLM}$ to be bijective or invertible, so,
\vspace*{-2mm}
\begin{equation}
 \scriptstyle
 M_{\mathcal{LLM}\rightarrow {}\mathcal{V}} =\mathcal{V}\bigl(x, y,\phi(x,y)\bigr)\circ \mathcal{LLM}\bigl(x, y,\phi(x,y)\bigr)\textsuperscript{-1}
\end{equation}
\vspace*{-4mm}
\begin{equation}
 \scriptstyle
 M_{\mathcal{V}\rightarrow {}\mathcal{LLM}} =\mathcal{LLM}\bigl(x, y,\phi(x,y)\bigr) \circ \mathcal{V}\bigl(x, y,\phi(x,y)\bigr)\textsuperscript{-1}
\end{equation}
 
Since a closed-form solution to analytically derive the mapping function $M_{\mathcal{V}\leftrightarrow{}\mathcal{LLM}}$ is not always feasible and since many task-based fine-tuning heads over a Transformer-based LLM involve fitting a linear classification layer, we propose a parameter-efficient linear-mapping technique {\tt Lin-Sem}. We estimate the mapping function within a empirical risk minimization framework by using a ridge regression between the two cross-modal representations. Mathematically,
\vspace*{-4mm}
\begin{multline}
  \scriptstyle
  M_{\mathcal{LLM}\rightarrow{}\mathcal{V}} \\ 
  \scriptstyle
  \leftarrow{} \text{minimize }  {((\mathcal{V}-\beta\mathcal{LLM})^T(\mathcal{V}-\beta\mathcal{LLM}) + \lambda\beta^T\beta)}
\end{multline}
\vspace*{-8mm}
\begin{multline}
  \scriptstyle
  M_{\mathcal{V}\rightarrow {}\mathcal{LLM}} \\
  \scriptstyle
  \leftarrow{} \text{minimize } {((\mathcal{LLM}-\beta\mathcal{V})^T(\mathcal{LLM}-\beta\mathcal{V}) + \lambda\beta^T\beta)}
\end{multline}
We assume \(\lambda\)=1 while \(\beta\) represents the  L\textsuperscript{2}-norm regularization parameter.

\paragraph{Datasets}

We evaluated our methods on the ECB+ \cite{cybulska2014using} and the AIDA Phase 1~\cite{tracey2022study} datasets. While the former is a popular, English-only CDCR benchmark containing a diverse range of news articles, the latter contains multimodal resources specific to Russia-Ukraine relations, in English, Russian, and Ukrainian. We focus only on the English documents.\footnote{The AIDA Phase 1 dataset was created for the DARPA Active Interpretation of Disparate Alternatives (AIDA) program and is available from the Linguistic Data Consortium (catalog number LDC2019E77). It is the only published ECR benchmark that contains multimodal resources specific to cross-document coreference. Events here are specifically in the domain of Russia-Ukraine relations and annotated based on both saliency and the potential for conflicting perspectives.}  For our experiments, we used training and evaluation splits following \citet{cybulska-vossen-2015-translating}  for ECB+ and \citet{tracey2022study} for AIDA Phase 1.  Table~\ref{tab:ecb_aida} shows corpus-level statistics for these two datasets.

\def\arraystretch{1}%
\begin{table}[t]
\centering
\resizebox{.48\textwidth}{!}{
    \begin{tabular}{ccccccc}  
    \toprule
     \multicolumn{1}{c}{~} & \multicolumn{3}{c}{ECB+} && \multicolumn{2}{c}{AIDA Phase 1}  \\
     \cmidrule{2-4} \cmidrule{6-7}
    \multicolumn{1}{c}{~} 
    Split & Train & Dev & Test && Practice & Eval \\ \midrule
	 
	Docs & 594 & 196 & 206 && 63 & 69  \\
	
	Event Mentions  & 3808 &  1245 & 1780 && 603 & 846\\

	Clusters & 1464 & 409 & 805 && 186 & 270\\
    
    Singletons & 1053 & 280 & 623 && 132 & 197  \\

    Images & 3808$^*$ &  1245$^*$ & 1780$^*$  && 417 & 662 \\
    
    \bottomrule
		
	\end{tabular}}
 \vspace*{-2mm}
  \caption[Corpora details: ECB+ and AIDA Phase 1]{ECB+ and AIDA corpus-level statistics. \citet{tracey2022study} refers to the provided train and test sets as ``practice'' and ``eval'', respectively.}
\label{tab:ecb_aida}
 \begin{tablenotes}
      \footnotesize
      \item $^*$Including images generated using Stable Diffusion.
    \end{tablenotes}
 \vspace*{-2mm}
\end{table}

\paragraph{Augmenting ECB+ with Images}

Since ECB+ does not provide images in their metadata, we scraped through the links provided in the documents and searched the Internet Archive for archived versions of articles with dead links. For original ECB documents without links, we manually search for keywords to retrieve articles. Out of 502 ECB+ document links, 43\% were broken, but 50\% could be recovered using {\tt web.archive.org}. Of 480 ECB documents, 51\% were located via Google search. We retrieved a total of 543 images; 235 of 982 documents had at least one associated image. 

In addition to the overall lack of images, the retrieved document-level images may be poor representatives of individual event mentions, leading to the sparsity problem mentioned in Sec.~\ref{sec:intro}. Therefore, we used Stable Diffusion~\cite{rombach2022high} to generate more relevant images and provide enough data to explore the contribution of multimodal information to ECR. Photo-realistic images were generated using sentences from ECB+ as prompts. Since a sentence can refer to multiple events, we provided an additional signal in the prompt by marking the event trigger with special tokens (<\texttt{m}> and </\texttt{m}>).

\paragraph{Image Encoding} 

To encode all images as vector representations, we used three variations of Vision Transformers (ViT; \citet{dosovitskiy2021image}, BEiT; \citet{bao2021BEiT}, and SWIN; \citet{liu2021swin}), as well as CLIP \cite{radford2021learning}. Resulting representations were the pooled output of the first-token representations from the last encoder layer for the image sequence, akin to the {\tt [CLS]} token in BERT variants. Encoding the images through distinct embedding spaces decoupled them from the original language inputs. 

\paragraph{Linear Projection Technique} 

To project image and text representations across modalities, we first created a concatenated 3,072D (768$\times$4) representation for an image/text pair. These concatenated representations contained the paired representation, the individual mention representations ({\tt Arg1} and {\tt Arg2}), and their element-wise product (in that order). Separate concatenated representations were constructed for each modality (see Fig.~\ref{fig:pipeline}).\footnote{All language representations came from the pretrained Longformer model~\cite{beltagy2020longformer}.} We then used a ridge regressor to calculate the linear coefficients by minimizing the squared distances between concatenated representations from each modality for the training set. This gave us two square (3,072$\times$3,072) ``bridge'' matrices: $M_{\mathcal{LLM}\rightarrow \mathcal{V}}$ and $M_{\mathcal{V}\rightarrow \mathcal{LLM}}$. We hypothesized that this bidirectional map retains crucial semantic information that a structure-preserving linear map would transfer between the two modalities. At evaluation, we matrix-multiplied the test concatenated representations with these matrices while maintaining the directionality of the linear map. These mapped representations were fed into a pairwise-scorer to get coreference clusters (see Fig.~\ref{fig:pipeline}).

\paragraph{Model Training and Fine-Tuning}

Following~\citet{humeau2020poly,cattan-etal-2021-cross-document}, {\it i.a.}, we trained separate pairwise scorers $P_{\theta, \theta'}$: $(AB, BA)$\textrightarrow $S_1, S_2$ on ECB+ and AIDA Phase 1. Here $AB$ and $BA$ are the 3,072D combined representations in A\textrightarrow B and B\textrightarrow A directions respectively, and $\theta$ and $\theta'$ are the parameters of the pairwise scorer and the LLM, respectively. This output two scores for each directional encoding, each representing the probability that the event mention pair was coreferent.\footnote{As a human reader would likely make a consistent coreference decision regardless of which event description she read first, we used the mean of the two scores as the final probability score for training and inference.}  Thereafter, we used the CoVal Scorer~\cite{moosavi-etal-2019-using} to form the final coreference clusters after applying transitive closure to identify the connected components with a threshold of 0.5 for all models. We used the same pairwise-scorer for all linear maps. 

\begin{figure}[h!]
  \begin{center}
      \includegraphics[width=0.3\textwidth]{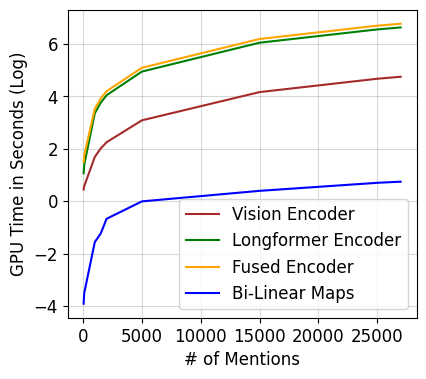}
  \end{center}
    \vspace*{-2mm}
\caption{Pairwise encoding time in GPU seconds (log-scale on y-axis) for text (Longformer), vision (ViT), and fused models vs. Bidirectional Linear Mapping ({\tt Lin-Sem}) as a function of the number of train pairs in ECB+.}
\label{fig:GPU-time}
 \vspace*{-2mm}
\end{figure}

For a direct multimodal comparison, we fine-tuned fused-modality models. We concatenated the image representations with the text representations and trained four separate pairwise scorers for each combination. Due to data sparsity of real images, we only trained fused models using generated event-centric images. Training took roughly 1.0 and 1.5 hours per epoch for the LLM and the fused models, respectively. For comparison, linear mapping took $\sim$3s to learn a mapping between modalities. Fig.~\ref{fig:GPU-time} shows log GPU seconds required for pairwise encoding for text, image, and fused modalities vs. bidirectional linear projection.

\subsection{Categorizing Mention Pair Difficulty}
\label{ssec:pair-diff}

\begin{figure}%
    \centering
    \includegraphics[width=.49\linewidth]{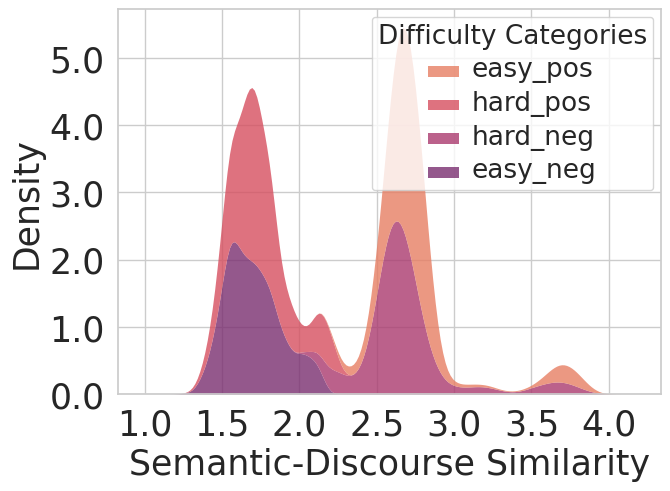}
    \includegraphics[width=.49\linewidth]{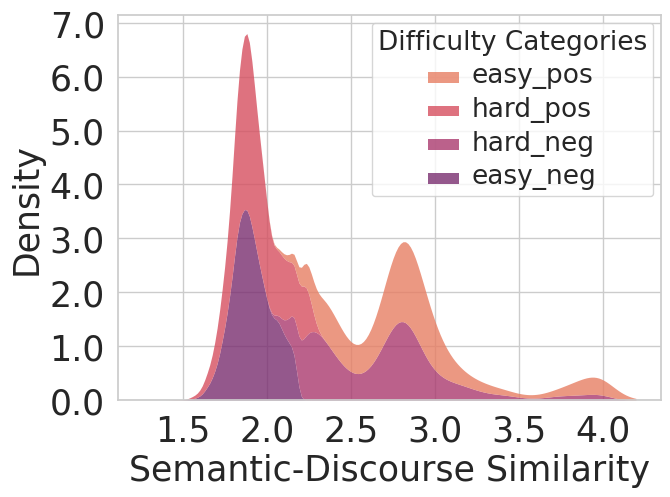}
    
 
 \vspace*{-2mm}
    \caption{Kernel Density Estimation plots of semantic-discourse similarity scores (including Wu-Palmer similarity) for mention pair difficulty categories in ECB+ (L) and AIDA Phase 1 (R), showing a clear demarcation of easy and hard pairs in positive and negative labels. {\tt easy\_pos} and {\tt hard\_neg} pairs have a high semantic similarity distribution while {\tt easy\_neg} and {\tt hard\_pos} pairs have lower semantic similarity distribution.}\label{fig:semantic_kdeplot}
 \vspace*{-2mm}
  
\end{figure}

To empirically evaluate the contribution of cross-modal information toward resolving challenging event mention pairs, we used the gold-standard coreference labels to categorize unseen pairs at inference as \textit{easy} or \textit{hard} based on semantic and discourse-level similarities. For semantic similarities, we use Wu-Palmer Similarity~\cite{wu1994verb}, and cosine similarity metrics. For discourse-level similarities, metadata in both datasets provides information about within-topic and within-document events which we used to score event similarities. For instance, an event pair within the same document and topic would get the highest discourse-level similarity score. These combined semantic and discourse similarity scores were then bucketed into {\it easy} and {\it hard} semantic transfer categories based on the means of coreferring and non-coreferring samples (see Fig.~\ref{fig:semantic_kdeplot}).

An example ``hard'' mention pair from ECB+, involving pronominal coreference, is (1) ``{\it In a move \textbf{<m>} \underline{that} \textbf{</m>} will expand its services division, Hewlett-Packard will acquire EYP Mission Critical Facilities}'' and (2) ``{\it HP to \textbf{<m>} \underline{Acquire} \textbf{</m>} Data Center Consultants.}''  This categorization allowed us to identify cases where multimodal features are distinctly useful based on proportion of correctly resolved {\it hard} pairs (see Sec.~\ref{sec:results}). Table~\ref{fig:easy_hard_samples} in Appendix~\ref{app:semantic_difficulty_category} shows examples of {\it easy} and {\it hard} pairs for coreferring and non-coreferring samples and their respective counts.

\paragraph{Computation of Semantic Difficulty Categories}
It is important to note that the ``hard'' and ``easy'' categories include both positive (coreferent) and negative (non-coreferent) samples. These categories are computed based on the assumption that easier coreferent (easy positive) samples should ideally have a higher overall similarity than harder ones, both in terms of semantics and at the topic and discourse level. Similarly, easier non-coreferent samples (easy negative) should ideally have a lower overall similarity. Hard coreferent (hard positive) pairs have lower overall similarity and hard non-coreferent (hard negative) pairs have higher overall similarity when compared to easy pairs of the same label.

Overall similarity for a given pair is computed as the sum of four individual scores:
\begin{enumerate}
    \item whether a pair comes from the same topic (1 for within-topic, 0 for not),
    \item whether a pair comes from the same document (1 for within-doc, 0 for not),
    \item the Wu-Palmer similarity of the trigger tokens in a pair, and
    \item the average cosine similarity of the vectors for the two sentences when encoded in both directions using the text-only, finetuned LLM (Longformer), inspired by \cite{ahmed20232}.
\end{enumerate}

For computing the cosine similarity scores, we take two mention-containing sentences \textit{A} and \textit{B} and cross-encode sentence \textit{A} in context before sentence \textit{B} and sentence \textit{B} in context after sentence \textit{A}. We then take the cosine similarity between these two encoded vectors. The positions of \textit{A} and \textit{B} are then reversed and they are again encoded with cross-attention in the same way. Because cross-attention is used, this results in different positional encodings for the two sentences and therefore a different cosine similarity value than the first calculation, so these values are then averaged for the final score.

Adding the aforementioned four scores gives us the final similarity scores for each pair in each label category (positive and negative). If the final similarity score for an individual positive pair is more than the mean final similarity score for all positive pairs, such a pair is categorized as easy positive. If it is less than this value, it is categorized as hard positive. On the other hand, if the final similarity score for an individual negative pair is more than the mean final similarity score for all negative pairs, the pair is categorized as hard negative, and if it is less than this value, it is categorized as easy negative.\footnote{The average final similarity for all positive samples over the ECB+ corpus is $2.25$, and the average final similarity for all negative samples is $2.14$. We assume AIDA Phase 1 comes from a disparate distribution, and so we categorize the difficulty of pairs in it independently using the same procedure.}  The plots in Fig. 2 show the differences in the distributions of different sample categories vs. the calculated similarity scores for both the corpora. See Appendix~\ref{app:semantic_difficulty_category} for more details with computed examples.

We use the gold coreference labels to obtain the label categories. However, since this categorization is only used as an evaluation tool for the initial round of experiments and then frozen for the ensembling experiments, the difficulty category-related information is never used during model training.

 \vspace*{-2mm}
\section{Results and Analysis}
\label{sec:results}
 \vspace*{-2mm}

We evaluate using established coreference metrics \cite{moosavi-etal-2019-using}, e.g., MUC~\cite{vilain1995model}, $B^{3}$~\cite{bagga1998algorithms}, $CEAF_e$, and CoNLL F1 (the average of MUC, $B^{3}$ and $CEAF_e$ F1) scores.

 \vspace*{-2mm}
\subsection{ECB+}
\label{ssec:results-ecb}
 \vspace*{-2mm}

We present results from \citet{held-etal-2021-focus} as a current, commonly accepted SOTA on ECB+, and from \citet{ahmed20232}, whose computationally-efficient pruning heuristic based on surface lemma similarity we follow to allow us to perform multiple experiments on a smaller compute budget.  Direct comparison to text-only model (\LLM) performance should be taken as a comparison to \citet{ahmed20232} due to the preprocessing. Table~\ref{tab:mm_cdcr_partial} shows detailed results.

\input{mm_cdcr_partial_ecb}

\paragraph{Text-only vs. Multimodal Models}
Despite the extra training time incurred in training a fused-modality model with concatenated features (see Fig.~\ref{fig:GPU-time}), we see that the performance of the fused multimodal models does not exceed that of the text-only model (Longformer using \citet{ahmed20232}'s preprocessing heuristic).  Interestingly, the performance gap between linearly-mapped systems and fused modality models is often quite small, despite the higher compute cost of training the fused model.  For instance, \LLM\textrightarrow \BEiT-gen and \LLM\textrightarrow \BEiT-real (Longformer embeddings mapped into \BEiT~space) slightly best the CoNLL F1 score of \BEiT-gen $\oplus$ \LLM, and \BEiT-real\textrightarrow \LLM~is only 0.5 F1 points lower.  Similar trends hold when comparing other fused modality models and their linearly-mapped counterparts, such as \LLM\textrightarrow \SWIN-gen, \LLM\textrightarrow \SWIN-real, and \SWIN-real\textrightarrow \LLM~ vs. \SWIN-gen $\oplus$ \LLM.


\paragraph{Semantic Transfer Categories}

In the coreferrence domain, one weakness of the CoNLL F1 metric is that specific evaluation metric-level details are obfuscated---this can be seen in Table~\ref{tab:prec_rec}: although the aforementioned examples achieve comparable CoNLL F1 scores, the linear mappings achieve a much higher MUC and $B^{3}$ recall, but lower precision, than the comparable fused models. Therefore, we do a proportional analysis of the correctly inferred (true positive) and misclassified (false positive and false negative) samples within the semantic transfer categories (see Table~\ref{tab:semantic_analysis}). These categorization labels were not used as supervision at any stage of training, fine-tuning, or mapping, and so an analysis of which models do better at which categories can illuminate different properties of the models, despite similar numerical performance. Table~\ref{tab:semantic_analysis} shows the proportion of each result category per model, of samples that would be considered ``hard'' according to the mention pair difficulty categorization described in Sec.~\ref{sec:method}.

\begin{table}[!htb]
\centering
\resizebox{\linewidth}{!}{
 \begin{tabular}{rrrrrr} 
\toprule
\multirow{2}{*}{Models} & \multicolumn{2}{c}{MUC} && \multicolumn{2}{c}{$B^3$} \\
        \cmidrule(lr){2-3} \cmidrule(lr){5-6}
& $R$ & $P$ && $R$ & $P$ \\
        \cmidrule(lr){1-6}

            \LLM \textrightarrow  \ViT-gen & 98.7 & 79.6 && 97.6 & 67.7 \\

       \LLM \textrightarrow  \BEiT-gen  & 99.1 & 79.6 && 97.9 & 67.7 \\

            \texttt{ViT}-gen $\oplus$ \texttt{LLM} & 80.9 & 89.7 && 85.4 & 86.9 \\

    \texttt{BEiT}-gen $\oplus$ \texttt{LLM} & 75.9 & 89.7 && 82.5 & 87.5 \\
 \bottomrule
 \end{tabular}}
 \vspace*{-2mm}
 \caption{MUC and $B^{3}$ precision and recall comparison between linear mappings and comparable fused models.}
 \label{tab:prec_rec}
\vspace*{-2mm}
\end{table}

Within true positives (TP), linearly-mapped models, using both real and generated images, tended to correctly retrieve a higher proportion of hard pairs compared to the text-only and fused models. For instance, for generated images, the hard sample proportion retrieved by text-to-image models is almost 4 percentage points higher than that of text-only or fused models, while image-to-text models, though lower on average, still also correctly retrieve a higher proportion of hard pairs. This effect appears slightly more pronounced on average in the case of real images (avg. 51.8\% hard pairs in TPs, compared to 50.1\% for generated images, and 46.6\% for text-only).


\begin{table}[htb]
\centering
\tiny
\resizebox{\columnwidth}{!}{
\begin{tabular}{*{4}{r}}
  \toprule
 \multicolumn{4}{c}{Semantic Transfer Categories} \\
   \cmidrule(lr){1-4}
 Models & TP-Hard &FP-Hard & FN-Hard \\ 
 \midrule
 \multicolumn{4}{c}{ECB+} \\
 \midrule
    \citet{ahmed20232} & 0.466 & 0.521 & 0.607 \\

    \cdashline{1-4}  

    \ViT-real\textrightarrow \LLM & 0.625 & 0.250 & 0.506 \\
    \BEiT-real\textrightarrow \LLM & 0.521 & 0.436 & 0.434 \\
    \SWIN-real\textrightarrow \LLM & 0.510 & 0.451 & 0.407 \\
    \CLIP-real\textrightarrow \LLM & 0.476 & 0.536 & 0.508 \\
    \LLM \textrightarrow \ViT-real & 0.507 & 0.456 & 0.441 \\
    \LLM \textrightarrow \BEiT-real & 0.506 & 0.000 & 0.000 \\
    \LLM \textrightarrow \SWIN-real & 0.496 & 0.438 & 0.700 \\
    \LLM \textrightarrow \CLIP-real & 0.505 & 0.452 & 0.708\\

    \cdashline{1-4}
    
    \texttt{ViT}-gen $\oplus$ \texttt{LLM} & 0.432 & 0.591 & 0.635 \\
    \texttt{BEiT}-gen $\oplus$ \texttt{LLM} & 0.437 & 0.606 & 0.584 \\
    \texttt{SWIN}-gen $\oplus$ \texttt{LLM} & 0.404 & 0.620 & 0.642 \\
    \texttt{CLIP}-gen $\oplus$ \texttt{LLM} & 0.477 & 0.506 & 0.729 \\
   
    \cdashline{1-4}
    
    \ViT-gen\textrightarrow \LLM & 0.487 & 0.472 & 0.521 \\
    \BEiT-gen\textrightarrow \LLM & 0.471 & 0.445 & 0.525 \\
    \SWIN-gen\textrightarrow \LLM  & 0.548 & 0.433 & 0.478 \\
    \CLIP-gen\textrightarrow \LLM  & 0.483 & 0.490 & 0.534 \\
    \LLM \textrightarrow \ViT-gen & 0.505 & 0.449 & 0.541 \\
    \LLM \textrightarrow \BEiT-gen & 0.506 & 0.451 & 0.000 \\
    \LLM \textrightarrow \SWIN-gen & 0.505 & 0.452 & 0.531 \\
    \LLM \textrightarrow \CLIP-gen & 0.506 & 0.451 & 0.632 \\
    \midrule
    \multicolumn{4}{c}{AIDA Phase 1} \\
    \midrule
    \LLM  & 0.561 & 0.385 & 0.695 \\
    \ViT-real\textrightarrow \LLM & 0.609 & 0.368 & 0.734 \\
    \BEiT-real\textrightarrow \LLM & 0.661 & 0.328 & 0.629 \\
    \SWIN-real\textrightarrow \LLM & 0.660 & 0.327 & 0.636 \\
    \CLIP-real\textrightarrow \LLM & 0.627 & 0.332 & 0.657 \\
    \LLM \textrightarrow \ViT-real & 0.643 & 0.346 & 0.929 \\
    \LLM \textrightarrow \BEiT-real & 0.638 & 0.352 & 0.749\\
    \LLM \textrightarrow \SWIN-real & 0.667 & 0.333 & 0.562 \\
    \LLM \textrightarrow \CLIP-real & 0.648 & 0.341 & 0.000\\

    \bottomrule
\end{tabular}}
 \vspace*{-2mm}
  \caption{Table showing the proportion of hard event pairs within the true positive (TP), false positive (FP) and false negative (FN) samples based on semantic transfer category (Sec.~\ref{sec:method}) for ECB+. Values of 0 indicate that no cases fit this category, resulting in zero numerator.}
    \label{tab:semantic_analysis}
 \vspace*{-2mm}
\end{table}

\paragraph{Ensembling Models}

The apparent facility of different models at correctly retrieving mention pairs of different semantic difficulties led to a question: since the mention pair difficulty was never used during training, fine-tuning, or mapping, and only as an analytic tool, could we split the mention pairs according to their difficulty, and use the different model types to handle mention pairs they on average appear to be better at?  We therefore built an ensembling approach using the text-only model to handle easier pairs, and performed a grid-search through different combinations of the previously-trained multimodal models to handle harder pairs. We allowed for different multimodal models to potentially handle hard-positive pairs and hard-negative pairs and used the combined results from all models to compute the coreference metrics.  Table~\ref{tab:mm_cdcr_ensemble_partial} shows the best performing ensembles.

\input{mm_cdcr_partial_ensemble_ecb}

Our best performing ensemble model used \ViT-real\textrightarrow \LLM~to handle hard negative pairs, \LLM \textrightarrow \BEiT-real, to handle hard positive pairs, and the text-only language model to handle easy pairs.  This resulted in a CoNLL F1 score of 91.9, with scores of 89.5 or higher across all components of MUC, $B^3$, or $CEAF_e$ metrics, showing the ability of this ensemble to score highly on, and balance, multiple measurements. Other ensembles, such as a variant that used \CLIP-real\textrightarrow \LLM~to handle hard negatives, performed at a similar level.  Two particularly interesting points emerge:
\begin{enumerate*}[label=\arabic*)]
    \item Using both real and generated images, \LLM \textrightarrow \BEiT~routinely performed best at handling hard positive pairs;
    \item Many ensemble models using {\tt Lin-Sem}, especially those using a $\mathcal{V}\rightarrow\mathcal{LLM}$ mapping for hard negatives and an $\mathcal{LLM}\rightarrow\mathcal{V}$ mapping for hard positives, outperform the fused model/text-only model ensembles, despite the simplicity of the linear transformation.
\end{enumerate*}
This suggests that not only can visual information be leveraged for correct coreference of semantically more difficult mention pairs, but also that visual information may contain fine-grained cues useful for splitting mention pairs while linguistic information is more useful to cluster them.

 \vspace*{-2mm}
\subsection{AIDA Phase 1}
\label{ssec:results-ldc}
 \vspace*{-2mm}

Table~\ref{tab:mm_cdcr_partial_ldc} presents a novel baseline on the multimodal AIDA Phase 1 data. This data contains unique challenges, such as a train set that is smaller than the test data, and event descriptions from sources with conflicting perspectives, explicitly addressing the ambiguity and perspective conflict challenges from Sec.~\ref{sec:intro}. Since this data comes with images mappable to individual event mentions, we evaluate using only the provided images.

As with ECB+, we find that models using linear mappings compete with or slightly outperform the text only model.  Using the same proportional analysis of correct and misclassified samples by difficulty category, we find that linearly-mapped models are also more likely than the text-only to resolve hard pairs correctly on this dataset (avg. hard pairs in TPs: 63.9\% for $\mathcal{V}\rightarrow\mathcal{LLM}$, 64.9\% for $\mathcal{LLM}\rightarrow\mathcal{V}$, and 56.1\% for text-only).

We then applied the same ensembling approach to the AIDA data, using the same combination of linear mappings and the LLM according to the difficulty of the mention pair.  Again we find that an ensemble model using a $\mathcal{V}\rightarrow\mathcal{LLM}$ mapping for hard negatives and an $\mathcal{LLM}\rightarrow\mathcal{V}$ mapping for hard positives performs best, although this time the model using \CLIP-real\textrightarrow \LLM~as the hard negative handler comes out on top.



\input{mm_cdcr_partial_ldc}

 \vspace*{-2mm}
\section{Discussion}
 \vspace*{-2mm}

Some specific example pairs where the text-only and fused models fail to link the pair, but ensembles correctly do so, expose certain features crucial for event coreference that are present in visual information and linearly transferable, but missing in text alone or scrambled during model fusion.
 
\paragraph{ECB+}

\begin{figure}[t]
  \centering
  \includegraphics[width=\columnwidth]{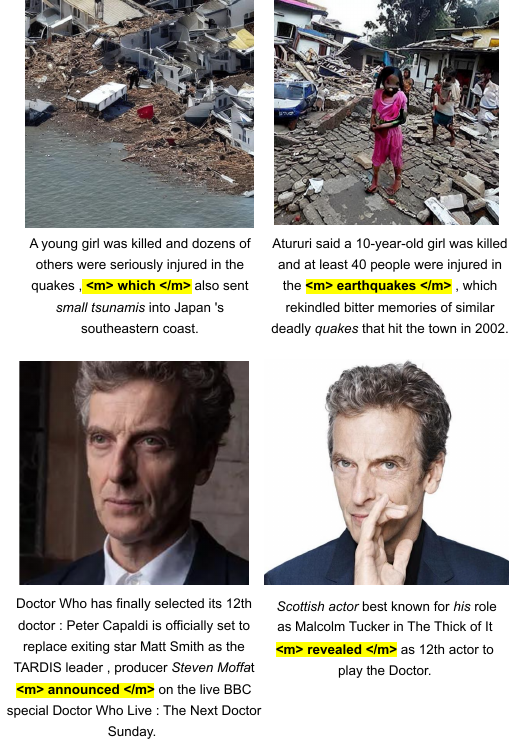}
 \vspace*{-2mm}
\caption{Sample coreferent event pairs from ECB+ that were correctly linked by our best multimodal ensemble (\ViT-real\textrightarrow \LLM~+ \LLM \textrightarrow \BEiT-real + \LLM), but not by the text-only model. Event-triggers are highlighted in yellow and text in italics illustrates lexical ambiguity or misleading lexical overlap.}  \label{fig:ecb_ensemble_examples}
 \vspace*{-2mm}
\end{figure}

ECB+ examples of this kind include event pairs that require some sense of visual grounding, temporal logic~\citep{schank1975scripts,ravi2023happens} or pronominal context to resolve. For instance, pairs with pronominal antecedents and misleading lexical overlap like ``{\it ...dozens of others were seriously injured in the quakes, \underline{which} also sent small tsunamis...}'' and ``{\it ...injured in the \underline{earthquakes} which rekindled bitter memories of similar deadly \textit{quakes}...}''\footnote{``[E]arthquakes'' vs. ``quakes'' is misleading lexical overlap as they refer to different earthquakes. The actual event triggers are underlined.} were missed by the LLM and fused models. Visual cues, such as damaged buildings or injured people (either in images generated using mentions as prompts, or already present in images in news articles) can help make the link. The aforementioned example is shown in Fig.~\ref{fig:ecb_ensemble_examples}, and the images are generated according to the ECB+ augmentation methodology (Sec.~\ref{sec:method}). Also in Fig.~\ref{fig:ecb_ensemble_examples}, {\it Steven Moffat} and {\it his} appear to be ambiguously overlapping to the text-only model, which missed the event mentions that are actually about Peter Capaldi. The two facial images, which are real images associated with the event mentions, help make the link. 

\paragraph{AIDA Phase 1}

Coreferent event mentions in the AIDA dataset are notable for conflicting information, and we find cases such as ``{\it Calling people tell about \underline{people that are} \underline{jumping out of the burning building}.}'' vs. ``{\it Forty-two people trapped by a fire on the third floor of the stately, Soviet-era Trades Unions building burned, \underline{suffocated} or jumped to their deaths.}'' Text-only fails to link ambiguous event triggers, but the images associated with each show the Trades Unions building in Odesa.  In such context-sensitive pairs, the paired visual representations (image domain {\tt Arg1} and {\tt Arg2} in Fig.~\ref{fig:pipeline}) in {\tt Lin-Sem} help resolve the coreference by capturing less ambiguous information from images while the text-only pairwise scorer found low contextual similarity between the event triggers.
Similarly, we see pairs with ambiguous context or pronominal anaphora, e.g., ``{\it Buzina, 45, \underline{was shot} dead}'' vs. ``{\it He was \underline{murdered}}'', are frequently missed by the LLM, but not by the ensemble models. In the case of this mention pair, both associated articles contain (different) pictures of the same individual, Oles Buzina, which, as with the ECB+ Peter Capaldi example, aids in the coreference\footnote{\citet{mcneely2022canonical} present strong evidence for the particular effectiveness of linear transformation in face recognition.}. Generally, for challenging corpora like AIDA Phase 1, we find visual features like faces, or background cues like angry protesters, press conferences, etc., act as cues for correctly resolving that pair. 

 \vspace*{-2mm}
\section{Conclusion}
\label{sec:conc}
 \vspace*{-2mm}

In this paper we have demonstrated the utility of multimodal information in cross-document event coreference. In particular, our results demonstrate that multimodal information is useful for resolving mention pairs whose triggers have low semantic and discourse-level similarity, rendering them difficult for text-only models. We developed a method ({\tt Lin-Sem}) for using linear transformations between embedding spaces to transfer semantic information between vision and language representation spaces, and used this technique in a model ensembling approach that used {\tt Lin-Sem} models to handle harder mention pairs and a text-only model for easier pairs.  We applied this approach to the popular ECB+ benchmark and established a novel baseline on the challenging, and explicitly multimodal, AIDA Phase 1 dataset \cite{tracey2022study}. Our best performing models beat text-only performance on these datasets by $\sim$3 F1 points and establish an upper bound on CDCR performance given the preprocessing used. Our ablation studies show that ensemble systems built upon our mention pair difficulty categories and using structure preserving linear maps can leverage event-specific visual cues to make correct coreference decisions about difficult mention pairs. These visual cues are of course absent in text only models, and are likely scrambled during standard multimodal fusion approaches. As such, our results present a strong case for the utility of multimodal information in NLU tasks like event coreference and argue for future increased development of such resources. Upon publication, we will release our processing pipeline and the generated/scraped images associated with ECB+.\footnote{The AIDA Phase 1 data must be properly obtained from the Linguistic Data Consortium.}

Our results should be considered in the context of our preprocessing assumptions.  We use a computationally-efficient pruning heuristic that allowed us to run the high volume of experiments we showcased on a lower compute budget, while demonstrating the utility of multimodal features for coreference. Our binary semantic transfer categories (\textit{easy}/\textit{hard}) do not currently account for semantic similarity between pairs that cross subtopics since corpora like the ECB+ corpus do not contain coreference annotations across sub-topics~\cite{bugert2021generalizing}. However, our framework can be easily expanded to corpora like FCC~\cite{bugert2020breaking}, with cross-subtopic events.

\vspace*{-2mm}
\section{Future Work}
\label{sec:future}
 \vspace*{-2mm}
Future directions in this line of research include exploring the feasibility of using multimodal cues to align/enhance representation spaces of monolingual LLMs, like the English-only Longformer, for Russian and Ukrainian mention pairs in the AIDA Phase 1 corpus. Given the efficiency of linear transformations and the rarity of coreference-specific parallel corpora, this may help alleviate the compute budgets needed for multilingual LLM pretraining for CDCR. Another interesting direction is evaluating our method for other challenging CDCR datasets like FCC~\cite{bugert2020breaking} which contains cross-subtopic events or the GVC~\cite{vossen2018don} where the SOTA is lower compared to benchmarks like ECB+. Lastly, this work represents a novel cross-modal case where affine transformations between embedding spaces has been shown to be useful (cf. \citet{mcneely2022canonical,nath2022phonetic,merullo2023linearly,ghaffari2023grounding}). Future work in this area entails a theoretical exploration of the properties of embedding spaces with a goal of finding performance guarantees where affine transformations successfully preserve information for different AI tasks.

\section*{Ethics Statement}

Our ablation studies required a non-trivial computation budget and concomitant resource usage, especially for the fused models with larger scoring heads on top of the LLM. Moreover, even though our {\tt Lin-Sem} framework is substantially compute-efficient, it still required cross-modal model encoding in generating representations for deploying our linear maps between them. The images generated for this task with diffusion models might reflect social, racial, or gender-based stereotypes as are commonly seen in large generative models. Due to the nature of the AIDA Phase 1 data's focus on Ukrainian-Russian conflict, the events described therein are likely to be distressing to some.




\section*{Acknowledgements}
This research was supported in part by grant award FA8750-18-2-0016 from the U.S. Defense Advanced Research Projects Agency (DARPA) to Colorado State University and the University of Colorado, and by a subcontract to the University of Colorado on grant award FA8750-19-2-1004 from DARPA. Views expressed herein do not reflect the policy or position of the Department of Defense or the U.S. Government. All errors are the responsibility of the authors.

%
\section*{Bibliographical References}
\label{reference}
\vspace*{-2mm}
\bibliographystyle{lrec-coling2024-natbib}
\bibliography{anthology,linear_map}
\appendix

\section{Examples of Semantic Difficulty Categories}
 \label{app:semantic_difficulty_category}
To illustrate how the combined scores of the four individual similarity scores are used to categorize the difficulty of mention pairs, consider an ECB+ cross-document event mention pair from each of the four semantic difficulty categories (Sec.~\ref{ssec:pair-diff}).

\textbf{Easy-positive (Easy-P)}
\begin{itemize}
    \item Sentence 1: Advanced Micro Devices to <m> \textit{acquire}</m> microserver vendor SeaMicro for \$334 million.
    \item Sentence 2: AMD to <m> \textit{Acquire} </m> Server Start - Up
\end{itemize}
In the above event mention pair, the final similarity score is $2.607 = 1 + 0 + 1 + 0.607$. Since this is more than the average of the final similarity for the positive samples ($2.25$), this mention pair is categorized as easy-positive.

\textbf{Hard-positive (Hard-P)}
\begin{itemize}
    \item Sentence 1: <m> \textit{4.6 quake} </m> rattles Sonoma County early Thursday.
    \item Sentence 2: <m> \textit{4-Plus Earthquake} </m> Recorded Near Healdsburg
\end{itemize}
In the above event mention pair, the final similarity score is $2.12 = 1 + 0 + 0.5 + 0.62$. Since this is less than the average of the final similarity for the positive samples ($2.25$), this mention pair is categorized as hard-positive.

\textbf{Easy-negative (Easy-N)}
\begin{itemize}
    \item Sentence 1: Apple <m> \textit{Unveils} </m> New Flagship Macbook Pro.
    \item Sentence 2: Next, global marketing VP Phil Schiller <m> \textit{announced} </m> updates to the MacBook line.
\end{itemize}
In the above event mention pair, the final similarity score is $1.92 = 1 + 0 + 0.22 + 0.70$. Since this is less than the average of the final similarity for the negative samples ($2.14$), this mention pair is categorized as easy-negative.

\textbf{Hard-negative (Hard-N)}
\begin{itemize}
    \item Sentence 1: <m> \textit{4.6 quake} </m> rattles Sonoma County early Thursday.
    \item Sentence 2: <m> \textit{4-Plus Earthquake} <m> Recorded Near Healdsburg
\end{itemize}
In the above event mention pair, the final similarity score is $2.60 = 1 + 0 + 1 + 0.60$. Since this is more than the average of the final similarity for the positive samples ($2.14$), this mention pair is categorized as hard-negative. Table~\ref{fig:easy_hard_samples} shows a few more examples of each type of semantic category and the total number of pairs in the ECB+ test set that fall into that category.

\begin{table*}[h]
\resizebox{\textwidth}{!}{
 \footnotesize  
    
    \begin{tabular}{l  p{5.6cm}  p{5.6cm}  c}
        \toprule
\textbf{Semantic Category} & \textbf{Pair1} & \textbf{Pair2} & \textbf{Pairs} \\ \midrule
 Easy-P & An Oklahoma man has pleaded not guilty to two first - degree murder charges for the {\tt <m>} \textbf{deaths} {\tt </m>} of an Arkansas woman and her fetus.'    & Oklahoma man pleads not guilty in {\tt <m>} \textbf{deaths} {\tt </m>} of Arkansas woman and her fetus. & 2661 \\\hline
Hard-P    & In a move {\tt <m>} \textbf{that} {\tt </m>} will expand its services division , Hewlett - Packard will acquire EYP Mission Critical Facilities , a New York company that offers data center consulting services.  & HP to {\tt <m>} \textbf{Acquire} {\tt </m>} Data Center Consultants & 2730 \\\hline
Easy-N     
& The UN has disputed claims that Hamas militants {\tt <m>} \textbf{fired} {\tt </m>} mortars from the Gaza school that has suffered the deadliest attack of the war with Israel. & Pressure to obtain a {\tt <m>} \textbf{ceasefire} {\tt </m>}  in Gaza has been mounting , with the EU warning Israel it was `` destroying '' its image , while Israeli forces on Tuesday ( 6 January ) killed at least 40 people during an attack on a United Nations-run school in Gaza. & 1315  \\\hline

Hard-N     
& Atururi said a 10-year-old girl was killed and at least 40 people were injured in the earthquakes , which rekindled bitter memories of similar deadly {\tt <m>} \textbf{quakes} {\tt </m>} that hit the town in 2002. & As aid started to arrive , hundreds of aftershocks continued to rattle the coastal city which was hit by the 7.6 and 7.5 magnitude {\tt <m>} \textbf{quakes} {\tt </m>} early on Sunday , cutting power and prompting a brief tsunami warning.  & 1082 \\\hline

\bottomrule
    \end{tabular}}
 \caption{Examples of easy and hard samples from our semantic transfer categories for the ECB+ test set. Suffixes P and N denote coreferring and non-coreferring pairs respectively according to the gold standard. The ``Pairs'' column shows the number of samples in that category.}
     \label{fig:easy_hard_samples}
\end{table*}

\section{Further Details on Definitions and Semantic Equivalence}
\label{app:semantic-eq-def}

Our definition of semantic equivalence is more generalized than those evaluated in works like~\citet{finch2005using} where semantic equivalence is specific to tasks like machine translation.

Although the approximate invertibilty of image representation functions like HOG~\cite{vondrick2013hoggles} has been proven, our assumption of bijectivity of both $\mathcal{V}$ and $\mathcal{LLM}$ (see Sec.~\ref{sec:method}, {\bf Semantic Equivalence}) is based on the fact that Transformer-based bidirectional encoders like the Longformer~\cite{beltagy2020longformer} are still arguably more bijective than static embeddings like word2vec~\cite{mikolov2013efficient} since the latter may have a many-to-one correspondence.

\section{Further Details on Image Generation and Vision Encoding}
 \label{app:image-generation}

To minimize GPU-compute requirements for image generation, we only generate one high-quality image for each event-mention in the ECB+ corpus. Our linear projection technique lets us encode events individually as well as pairwise without expending additional resources for generating image representations of paired mentions. 
 
For generating high-quality, photo-realistic images we chose a guidance scale of 7.5. This enabled us to generate images that were both creative and relevant to coreference-specific natural language descriptions. In order to ensure a more balanced trade-off between efficiency and quality, we set the number of inference steps to 15. The resulting RBG images had a resolution of 512$\times$512. The image generation process for the entire ECB+ corpus (6,833 event mentions) took $\sim$4 hours on an NVIDIA GeForce RTX 3090. 

To obtain image embeddings, we used the four models mentioned (ViT, BEiT, SWIN, and CLIP) from HuggingFace.\footnote{\url{https://huggingface.co}} We use the generated images to get both individual as well as paired-image embeddings from each of the Transformer-based models (excluding CLIP) after converting the original image to 224$\times$224 resolution using the AutoImageProcessor on HuggingFace. CLIP requires an additional linguistic supervision component along with image inputs. As such, we used sentences containing the respective event mentions while ensuring a maximum token length of 77 around the event mentions to retain context. More specifically, these embeddings are the pooled output of the classification or first-token representations from the last encoder layer for the image sequence. 

\section{Pairwise Scorer Hyperparameters}
\label{sec:scorer_hyperparameters}

Our supervised pairwise scorer trained on top of the LLM is a two-layer (768 and 128 neurons) neural network trained along with the base Longformer~\cite{beltagy2020longformer}, with binary cross entropy (BCE) loss and final sigmoid activation for 10 epochs over the annotated (gold) labels. We add two special tokens ({\tt <m>} and {\tt </m>}) to the LLM vocabulary while training the scorer, that trigger a specific event mention while encoding the text. This helps us generate a contextualized representation of the text akin to the {\tt [SEP]} token in BERT~\cite{devlin2018bert}. We use the Adam optimizer~\cite{kingma2014adam} with a learning rate of $1e-4$ for the scorer and $1e-5$ for the LLM.

We use the same hyperparameters mentioned above for our fused models, except for the input layer size (6,144 = 3,072$\times$2) of the pairwise scorers. For training both the LLM and the fused models, we utilize a single NVIDIA A100 GPU with 80GB memory.

\section{Impact of Coreference Evaluation Metrics}

For coreference tasks, the metric chosen has a substantial impact on the numerical results.  MUC is a link-based metric that calculates the minimum number of \textit{missing} links between mentions in the predicted entities in comparison to the gold-standard entities to get its recall and precision. It is also the least discriminative since it does not differentiate whether an extra link merges two singletons. $B^{3}$ is a mention-based metric that calculates overall recall and precision based on a combination of the recall and precision of the individual mentions. Unlike MUC, the presence of singletons in the corpus disproportionately affects $B^{3}$ scores. $CEAF_e$ intends to overcome $B^{3}$'s tendency to use a mention more than once when comparing entities containing that mention. It uses an optimized mapping to align the entities in key and response to calculate its precision and recall. This may explain why the precision and recall trends in our results are inverted when using $B^{3}$/MUC vs. when using $CEAF_e$.
\input{mm_cdcr_ensembletable}
\input{mm_cdcr_fulltable_ldc}

Table~\ref{tab:mm_cdcr_ensemble} and Table~\ref{tab:mm_cdcr_full_ldc} shows detailed results including the precision and recall scores from various metrics for the ECB+ and AIDA Phase 1 corpus respectively. Empirically, the nature and the extent of such semantic transfer can also be analyzed via the MUC metric, in which text to image models performed closest to the text-only LLM baseline \citet{ahmed20232} while outperforming fused models for the ECB+ corpus. Here, linear semantic transfer reduces the missing links generated after partitioning for calculating recall. Because the MUC is link-based, it calculates missing links needed to replicate the gold cluster chains. As such, fine-grained semantic transfer using linear maps leads to a high recall system with equivalent F1 scores compared to the text-only LLM even when precision takes a hit. This trend is also seen in $B^{3}$. The remaining metric, $CEAF_e$, shows an inverse trend where recall for text to image models is lower than the text-only LLM as well as fused models. However, lower recall is balanced by a increase in precision for text to image models relative to the others. In general, some semantic transfer is observed in all the three metrics albeit with some variations. An in-depth study of how the extent of semantic information transfer varies between metrics is a part of future work. 

\section{On CLIP as Image-Generator}
We include CLIP \cite{radford2021learning} as one of the image encoders to explore the extent of linear semantic information transfer when a vision encoder is trained with linguistic supervision (unlike the other three vision encoders chosen). This allows us to conduct more exhaustive ablation experiments to study the fidelity of such multimodal semantic transfer. 

Although the latent diffusion model used for image generation also leverages a CLIP-guided image synthesis, there is no apparent information overlap between the latent spaces of the former with the CLIP encoder due to the nature of the event coreference task as well as the separate training objectives between the two. For instance, an event-coreference sample sentence may contain more than one event while captions usually contain one major event (or actions). Moreover, captions used to train CLIP are on average more precise and shorter sequences compared to event-coreference documents which include language not directly relevant for inferring that specific coreference. As such, our design choice to include CLIP as one of the vision-encoders still satisfies the requirement for the generalizability of such models for linear semantic transfer without giving it an undue advantage through overlap of the CLIP text encoder and the image generation model.

\end{document}

%% file: mm_cdcr_partial_ecb.tex
\begin{table}[!htb]
\centering
\resizebox{\linewidth}{!}{
 \begin{tabular}{rrrrr} 
\toprule
Models & MUC & $B^3$ & $CEAFe$ & CoNLL\\
        \cmidrule(lr){1-5}
   
        \citet{held-etal-2021-focus} & 87.5 & 86.6 & 82.9 & 85.7 \\

    \citet{ahmed20232}   &\textbf{90.8} & \textbf{86.7} & \textbf{84.7} & \textbf{87.4} \\
        \cdashline{1-5}
 
    \ViT-real\textrightarrow \LLM  & 6.9 & 63.1 & 55.1 &  41.7 \\

     \BEiT-real\textrightarrow \LLM  & 87.3 & 80.3 & 76.7 & 81.4  \\

    \SWIN-real\textrightarrow \LLM  & 87.6 & 79.7 & 76.5 & 81.3 \\

    \CLIP-real\textrightarrow \LLM  & 24.7 & 66.3 & 57.5 & 49.5 \\

     \LLM \textrightarrow  \ViT-real & 88.2 & 80.1 & 77.5 & 81.9 \\

    \LLM \textrightarrow  \BEiT-real  & 88.3 & 80.0 & 77.4 & 81.9 \\
        
     \LLM \textrightarrow  \SWIN-real   & 87.9 & 80.3 & 77.8 & 82.0 \\

    \LLM \textrightarrow  \CLIP-real  & 88.3 & 80.0 & 77.4 & 81.9 \\
       
        \cdashline{1-5}
     \texttt{ViT}-gen $\oplus$ \texttt{LLM} & 85.1 & 86.1 & 80.7 & 84.0 \\

    \texttt{BEiT}-gen $\oplus$ \texttt{LLM} & 82.2 & 84.9 & 78.1 & 81.7 \\

    \texttt{SWIN}-gen $\oplus$ \texttt{LLM}  & 82.5 & 85.1 & 78.7 & 82.1 \\

    \texttt{CLIP}-gen $\oplus$ \texttt{LLM} & 89.3 & 84.2 & 82.6 & 85.4 \\
        
        \cdashline{1-5}

     \ViT-gen\textrightarrow \LLM & 77.4 & 78.8 & 71.5 & 75.9 \\

    \BEiT-gen\textrightarrow \LLM  & 77.8 & 79.8 & 73.7 & 77.1 \\

    \SWIN-gen\textrightarrow \LLM  & 79.5 & 79.6 & 73.4 &  77.5 \\

    \CLIP-gen\textrightarrow \LLM  & 83.0 & 82.1 & 76.3 &  80.5 \\

    \LLM \textrightarrow  \ViT-gen & 88.1 & 80.0 & 77.2 & 81.8 \\

    \LLM \textrightarrow  \BEiT-gen  & 88.3 & 80.0 & 77.4 & 81.9 \\

    \LLM \textrightarrow  \SWIN-gen   & 88.2 & 80.1 & 77.4 &  81.9 \\

    \LLM \textrightarrow  \CLIP-gen  & 88.3 & 80.0 & 77.4 & 81.9 \\
 \bottomrule
 \end{tabular}}
 \vspace*{-2mm}
 \caption{MM-CDCR  F1 scores for MUC, $B^3$, $CEAFe$ and CoNLL on ECB+ test set, using LLM only, {\tt Lin-Sem} (``\textrightarrow''), and domain-fused finetuned versions (``$\oplus$''). Cited works are previous benchmarks on text-only CDCR. {\bf Bold} indicates the best performer on each metric. ``-real'' indicates that the vision space was encoded with real images, while ``-gen'' indicates generated images.}
 \label{tab:mm_cdcr_partial}
\vspace*{-2mm}
\end{table}

%% file: mm_cdcr_partial_ensemble_ecb.tex
\begin{table}[!htb]
\centering
\resizebox{\linewidth}{!}{
 \begin{tabular}{rrrrr} 
\toprule
Models & MUC & $B^3$ & $CEAFe$ & CoNLL\\
        \cmidrule(lr){1-5}
        \citet{held-etal-2021-focus} & 87.5 & 86.6 & 82.9 & 85.7 \\

    \citet{ahmed20232}   & 90.8 & 86.7 & 84.7 & 87.4 \\
        \cdashline{1-5}
       \texttt{ViT}-gen $\oplus$ \texttt{LLM + LLM} & 89.1 & 86.5 & 84.8 &  86.8   \\

       \texttt{BEiT}-gen $\oplus$ \texttt{LLM + LLM} & 87.5 & 85.7 & 83.9 &  85.7 \\

        \texttt{SWIN}-gen $\oplus$ \texttt{LLM + LLM}  & 87.5 & 85.9 & 83.8 & 85.7 \\

        \texttt{CLIP}-gen $\oplus$ \texttt{LLM + LLM} & 90.1 & 85.3 & 83.8 & 86.4 \\
        
        \cdashline{1-5}
    \ViT-gen\textrightarrow \LLM~+ \LLM \textrightarrow \BEiT-gen + \LLM & 90.8 & 85.2 & 84.8  & 86.9  \\

      \BEiT-gen\textrightarrow \LLM~+ \LLM \textrightarrow \BEiT-gen + \LLM & 91.3 & 85.5 & 86.5 & 87.8\\

       \SWIN-gen\textrightarrow \LLM~+ \LLM \textrightarrow \BEiT-gen + \LLM & 90.4 & 84.4 & 83.8 & 86.2 \\

     \CLIP-gen\textrightarrow \LLM~+ \LLM \textrightarrow \BEiT-gen + \LLM & 91.2 & 85.3 & 85.7 & 87.4 \\

     \LLM \textrightarrow \ViT-gen + \LLM \textrightarrow \BEiT-gen + \LLM & 88.7 & 82.3 & 79.4 & 83.5 \\

    \LLM \textrightarrow \BEiT-gen + \LLM   & 88.7 & 82.2 & 79.1 &  83.3 \\

     \LLM \textrightarrow \SWIN-gen + \LLM \textrightarrow \BEiT-gen + \LLM   & 88.7 & 82.2 & 79.1 & 83.3 \\

     \LLM \textrightarrow \CLIP-gen + \LLM \textrightarrow \BEiT-gen + \LLM  & 88.7 & 82.2 & 79.1 & 83.3  \\
        
        \cdashline{1-5}
        
     \ViT-real\textrightarrow \LLM~+ \LLM \textrightarrow \BEiT-real + \LLM & \textbf{94.5} & \textbf{89.5} & \textbf{91.8} & \textbf{91.9} \\

     \BEiT-real\textrightarrow \LLM~+ \LLM \textrightarrow \BEiT-real + \LLM & 88.9 & 82.4 & 79.7 & 83.7\\

     \SWIN-real\textrightarrow \LLM~+ \LLM \textrightarrow \BEiT-real + \LLM & 88.7 & 82.2 & 79.1 &  83.3 \\

     \CLIP-real\textrightarrow \LLM~+ \LLM \textrightarrow \BEiT-real + \LLM & 94.3 & 89.3 & 91.6 & 91.7 \\

    \LLM \textrightarrow \ViT-real + \LLM \textrightarrow \BEiT-real + \LLM & 88.7 & 82.3 & 79.3 & 83.4 \\

     \LLM \textrightarrow \BEiT-real + \LLM  & 88.7 & 82.2 & 79.1 & 83.3 \\

     \LLM \textrightarrow \SWIN-real + \LLM \textrightarrow \BEiT-real + \LLM   & 89.0 & 82.7 & 80.1 & 83.9 \\

    \LLM \textrightarrow \CLIP-real + \LLM \textrightarrow \BEiT-real + \LLM  & 88.7 & 82.2 & 79.1 &  83.3   \\

 \bottomrule
 \end{tabular}}
 \vspace*{-2mm}
 \caption{MM-CDCR MUC, $B^3$, $CEAFe$ and CoNLL F1 results on ECB+ test set, using ensemble models. Format follows Table~\ref{tab:mm_cdcr_partial}. Ensemble model names follow the format {\it Hard-N model + Hard-P model + Easy pairs model}. \LLM~was always used to handle Easy pairs. The best performing models for hard negative and hard positives were found using a grid search through different combinations of multimodal models. If only one model besides \LLM~is listed, that model was used to handle all Hard pairs.}
\label{tab:mm_cdcr_ensemble_partial}
\vspace*{-2mm}
\end{table}

%% file: mm_cdcr_partial_ldc.tex
\begin{table}[!htb]
\centering
\resizebox{\linewidth}{!}{
 \begin{tabular}{rrrrr} 
\toprule
Models & MUC & $B^3$ & $CEAFe$ & CoNLL\\
        \cmidrule(lr){1-5}
        \LLM   & 80.7 & 49.5 & 54.1 & 61.4 \\
        \cdashline{1-5}

        ViT-real\textrightarrow \LLM  & 85.9 & 38.4 & 52.7 & 59.0  \\

        \BEiT-real\textrightarrow \LLM  & 85.7 & 42.6 & 57.9 & 62.1  \\

        \SWIN-real\textrightarrow \LLM  & 82.9 & 46.4 & 55.8 & 61.7  \\

        \CLIP-real\textrightarrow \LLM  & 78.5 & \textbf{52.4} & 53.5 & 61.5 \\

        \LLM \textrightarrow  \ViT-real & 86.3 & 37.3 & 52.7 & 58.8  \\

        \LLM \textrightarrow  \BEiT-real  & 85.7 & 40.2 & 53.1 & 59.7  \\
        
        \LLM \textrightarrow  \SWIN-real    & 86.2 & 39.1 & 54.4 & 59.9 \\

        \LLM \textrightarrow  \CLIP-real  & 86.2 & 37.1 & 52.3 & 58.5  \\
 
        \cdashline{1-5}

        \ViT-real\textrightarrow \LLM~+ \LLM \textrightarrow \BEiT-real + \LLM & 86.2 & 39.6 & 54.4 & 60.1 \\

        \BEiT-real\textrightarrow \LLM~+ \LLM \textrightarrow \BEiT-real + \LLM & \textbf{87.1} & 42.1 & 60.4 & 63.2 \\

         \SWIN-real\textrightarrow \LLM~+ \LLM \textrightarrow \BEiT-real + \LLM & \textbf{87.1} & 42.5 & 60.5 & 63.4 \\

        \CLIP-real\textrightarrow \LLM~+ \LLM \textrightarrow \BEiT-real + \LLM & \textbf{87.1} & 43.8 & \textbf{62.8} & \textbf{64.6}\\

        \LLM \textrightarrow \ViT-real + \LLM \textrightarrow \BEiT-real + \LLM & 86.2 & 39.0 & 53.5 & 59.6 \\

        \LLM \textrightarrow \BEiT-real + \LLM & 85.8 & 40.8  & 54.1 & 60.2  \\

        \LLM \textrightarrow \SWIN-real + \LLM \textrightarrow \BEiT-real + \LLM   & 86.6 & 40.7 & 56.6 & 61.3 \\

        \LLM \textrightarrow \CLIP-real + \LLM \textrightarrow \BEiT-real + \LLM  & 86.2 & 39.0 & 53.5 & 59.6   \\ 

 \bottomrule
 \end{tabular}}
 \vspace*{-2mm}
 \caption{MM-CDCR MUC, $B^3$, $CEAFe$ and CoNLL F1 results on AIDA Phase 1 Eval set. Format follows Tables~\ref{tab:mm_cdcr_partial} \&~\ref{tab:mm_cdcr_ensemble_partial}. \LLM~denotes Longformer evaluated with \citet{ahmed20232}'s methodology.}
    \label{tab:mm_cdcr_partial_ldc}

\vspace*{-2mm}
\end{table}

%% file: mm_cdcr_ensembletable.tex
\begin{table*}[!ht]
    \centering
    \tiny
    \resizebox{\textwidth}{!}{
    \begin{tabular}{@{}lrrrrrrrrrrrrrrrrrrr@{}}
    \toprule
    & \multirow{2}{*}{Models} && \multicolumn{3}{c}{MUC} && \multicolumn{3}{@{}c@{}}{$B^3$} & & \multicolumn{3}{c}{$CEAFe$} && CoNLL\\
    \cmidrule{4-6} \cmidrule{8-10} \cmidrule{12-14} \cmidrule{16-16}
    &&& $R$ & $P$ & $F_1$ && $R$ & $P$ & $F_1$ && $R$ & $P$ & $F_1$ && \multicolumn{1}{r}{$F_1$}  \\ 
   \midrule
        & \texttt{ViT}-gen $\oplus$ \texttt{LLM + LLM} && 89.2  & 89.0 & 89.1 && 90.5 & 82.9 & 86.5 && 84.7 & 84.9 & 84.8 &&  86.8   \\

        & \texttt{BEiT}-gen $\oplus$ \texttt{LLM + LLM} && 86.5  & 88.6 & 87.5 && 88.5 & 83.0 & 85.7 && 85.2 & 82.7 & 83.9 &&  85.7 \\

        & \texttt{SWIN}-gen $\oplus$ \texttt{LLM + LLM}  && 85.7  & 89.3 & 87.5 && 88.0 & 83.8 & 85.9 && 85.8 & 81.8 & 83.8 && 85.7 \\

        & \texttt{CLIP}-gen $\oplus$ \texttt{LLM + LLM} && 94.8  & 85.9 & 90.1 && 95.0 & 77.3 & 85.3 && 78.5 & 89.8 & 83.8 && 86.4 \\
        
        \cdashline{1-20}
        & \ViT-gen\textrightarrow \LLM~+ \LLM \textrightarrow \BEiT-gen + \LLM && 96.4  & 85.8 & 90.8 && 96.0 & 76.7 & 85.2 && 78.4 & 92.2 & 84.8  && 86.9  \\

        & \BEiT-gen\textrightarrow \LLM~+ \LLM \textrightarrow \BEiT-gen + \LLM && 96.0  & 87.1 & 91.3 && 95.6 & 77.4 & 85.5 && 81.2 & 92.7 & 86.5 && 87.8\\

        & \SWIN-gen\textrightarrow \LLM~+ \LLM \textrightarrow \BEiT-gen + \LLM && 96.6  & 84.9 & 90.4 && 96.1 & 75.2 & 84.4 && 76.8 & 92.2 & 83.8 && 86.2 \\

        & \CLIP-gen\textrightarrow \LLM~+ \LLM \textrightarrow \BEiT-gen + \LLM && 96.7  & 86.4 & 91.2 && 96.1 & 76.7 & 85.3 && 79.4 & 92.9 & 85.7 &&  87.4 \\

        & \LLM \textrightarrow \ViT-gen + \LLM \textrightarrow \BEiT-gen + \LLM && 96.9  & 81.7 & 88.7 && 96.3 & 71.8 & 82.3 && 70.4 & 90.9 & 79.4 &&  83.5 \\

        & \LLM \textrightarrow \BEiT-gen + \LLM   && \textbf{97.0}  & 81.6 & 88.7 && \textbf{96.4} & 71.6 & 82.2 && 70.1 & 90.8 & 79.1 &&  83.3 \\

        & \LLM \textrightarrow \SWIN-gen + \LLM \textrightarrow \BEiT-gen + \LLM   && \textbf{97.0}  & 81.6 & 88.7 && \textbf{96.4} & 71.6 & 82.2 && 70.1 & 90.8 & 79.1 && 83.3 \\

        & \LLM \textrightarrow \CLIP-gen + \LLM \textrightarrow \BEiT-gen + \LLM  && \textbf{97.0}  & 81.6 & 88.7 && \textbf{96.4} & 71.6 & 82.2 && 70.1 & 90.8 & 79.1 && 83.3  \\
        
        \cdashline{1-20}
        
        & \ViT-real\textrightarrow \LLM~+ \LLM \textrightarrow \BEiT-real + \LLM && 95.9  & \textbf{93.2} & \textbf{94.5} && 95.6 & \textbf{84.1} & \textbf{89.5} && \textbf{90.2} & 93.4 & \textbf{91.8} && \textbf{91.9} \\

        & \BEiT-real\textrightarrow \LLM~+ \LLM \textrightarrow \BEiT-real + \LLM && 96.9  & 82.0 & 88.9 && 96.3 & 72.0 & 82.4 && 71.0 & 91.0 & 79.7 &&  83.7\\

         & \SWIN-real\textrightarrow \LLM~+ \LLM \textrightarrow \BEiT-real + \LLM && \textbf{97.0}  & 81.6 & 88.7 && \textbf{96.4} & 71.6 & 82.2 && 70.1 & 90.8 & 79.1 &&  83.3 \\

        & \CLIP-real\textrightarrow \LLM~+ \LLM \textrightarrow \BEiT-real + \LLM && 95.9  & 92.9 & 94.3 && 95.6 & 83.8 & 89.3 && 89.8 & \textbf{93.5} & 91.6 &&  91.7 \\

        & \LLM \textrightarrow \ViT-real + \LLM \textrightarrow \BEiT-real + \LLM && \textbf{97.0}  & 81.8 & 88.7 && \textbf{96.4} & 71.7 & 82.3 && 70.3 & 90.9 & 79.3 && 83.4 \\

        & \LLM \textrightarrow \BEiT-real + \LLM  && \textbf{97.0}  & 81.6 & 88.7 && \textbf{96.4} & 71.6 & 82.2 && 70.1 & 90.8 & 79.1 && 83.3 \\

        & \LLM \textrightarrow \SWIN-real + \LLM \textrightarrow \BEiT-real + \LLM   && 96.9  & 82.2 & 89.0 && 96.3 & 72.4 & 82.7 && 71.4 & 91.1 & 80.1 && 83.9 \\

        & \LLM \textrightarrow \CLIP-real + \LLM \textrightarrow \BEiT-real + \LLM  && \textbf{97.0}  & 81.6 & 88.7 && \textbf{96.4} & 71.6 & 82.2 && 70.1 & 90.8 & 79.1 &&  83.3   \\

    \bottomrule
    \end{tabular}}
    \caption{MM-CDCR MUC, $B^3$, $CEAFe$ and CoNLL F1 results on ECB+ test set, using ensemble models. Format follows Table~\ref{tab:mm_cdcr_partial}. Ensemble model names follow the format {\tt Hard-N model + Hard-P model + Easy pairs model}. \LLM~was always used to handle Easy pairs. The best performing models for hard negative and hard positives were found using a grid search through different combinations of multimodal models. Where only one model is besides \LLM~is listed, that model was used to handle all Hard pairs.}
    \label{tab:mm_cdcr_ensemble}
\end{table*}

%% file: mm_cdcr_fulltable_ldc.tex
\begin{table*}[!ht]
    \centering
    \tiny
    \resizebox{\textwidth}{!}{
    \begin{tabular}{@{}lrrrrrrrrrrrrrrrrrrr@{}}
    \toprule
    & \multirow{2}{*}{Models} && \multicolumn{3}{c}{MUC} && \multicolumn{3}{@{}c@{}}{$B^3$} & & \multicolumn{3}{c}{$CEAFe$} && CoNLL\\
    \cmidrule{4-6} \cmidrule{8-10} \cmidrule{12-14} \cmidrule{16-16}
    &&& $R$ & $P$ & $F_1$ && $R$ & $P$ & $F_1$ && $R$ & $P$ & $F_1$ && \multicolumn{1}{r}{$F_1$}  \\ 
   \midrule
        & \LLM   &&  83.3  & 78.2 & 80.7 && 81.6 & 35.6 & 49.5 && 50.3 & 58.6 & 54.1 && 61.4 \\
        \cdashline{1-20}

        & \ViT-real\textrightarrow \LLM  && 96.4  & 77.5 & 85.9 && 96.3 & 24.0 & 38.4 && 39.1 & 81.1 & 52.7 && 59.0  \\

        & \BEiT-real\textrightarrow \LLM  && 93.8  & 78.9 & 85.7 && 93.7 & 27.5 & 42.6 && 46.4 & 77.3 & 57.9 && 62.1  \\

        & \SWIN-real\textrightarrow \LLM  && 87.7  & 78.5 & 82.9 && 86.5 & 31.7 & 46.4 && 48.9 & 65.0 & 55.8 && 61.7  \\

        & \CLIP-real\textrightarrow \LLM  && 78.5  & 78.6 & 78.5 && 78.3 & \textbf{39.3} & \textbf{52.4} && \textbf{53.6} & 53.4 & 53.5 && 61.5 \\

        & \LLM \textrightarrow  \ViT-real && \textbf{97.4}  & 77.5 & 86.3 && 97.2 & 23.1 & 37.3 && 38.3 & \textbf{84.7} & 52.7 && 58.8  \\

        & \LLM \textrightarrow  \BEiT-real  && 95.7  & 77.6 & 85.7 && 95.3 & 25.5 & 40.2 && 39.9 & 79.2 & 53.1 && 59.7  \\
        
        & \LLM \textrightarrow  \SWIN-real    && 96.2  & 78.0 & 86.2 && 95.9 & 24.5 & 39.1 && 40.9 & 81.2 & 54.4 && 59.9 \\

        & \LLM \textrightarrow  \CLIP-real  && \textbf{97.4}  & 77.4 & 86.2 && \textbf{97.2} & 22.9 & 37.1 && 37.9 & 84.6 & 52.3 && 58.5  \\




        
        \cdashline{1-20}









         & \ViT-real\textrightarrow \LLM~+ \LLM \textrightarrow \BEiT-real + \LLM && 96.5  & 77.9 & 86.2 && 96.4 & 24.9 & 39.6 && 40.5 & 82.9 & 54.4 && 60.1 \\

        & \BEiT-real\textrightarrow \LLM~+ \LLM \textrightarrow \BEiT-real + \LLM && 96.2  & 79.6 & \textbf{87.1} && 96.0 & 27.0 & 42.1 && 47.0 & 84.5 & 60.4 && 63.2 \\

         & \SWIN-real\textrightarrow \LLM~+ \LLM \textrightarrow \BEiT-real + \LLM && 96.0  & 79.7 & \textbf{87.1} && 95.9 & 27.3 & 42.5 && 47.3 & 84.0 & 60.5 && 63.4 \\

        & \CLIP-real\textrightarrow \LLM~+ \LLM \textrightarrow \BEiT-real + \LLM && 95.3  & \textbf{80.3} & \textbf{87.1} && 95.4 & 28.4 & 43.8 && 50.2 & 83.7 & \textbf{62.8} && \textbf{64.6}\\

        & \LLM \textrightarrow \ViT-real + \LLM \textrightarrow \BEiT-real + \LLM && 96.9  & 77.7 & 86.2 && 96.7 & 24.4 & 39.0 && 39.4 & 83.1 & 53.5 && 59.6 \\

        & \LLM \textrightarrow \BEiT-real + \LLM && 95.5  & 77.9 & 85.8 && 95.2 & 26.0 & 40.8 && 41.1 & 79.2 & 54.1 && 60.2  \\

        & \LLM \textrightarrow \SWIN-real + \LLM \textrightarrow \BEiT-real + \LLM   && 96.4  & 78.6 & 86.6 && 96.1 & 25.8 & 40.7 && 43.0 & 82.9 & 56.6 && 61.3 \\

        & \LLM \textrightarrow \CLIP-real + \LLM \textrightarrow \BEiT-real + \LLM  && 96.9  & 77.7 & 86.2 && 96.7 & 24.4 & 39.0 && 39.4 & 83.1 & 53.5 && 59.6   \\

    \bottomrule
    \end{tabular}}
    \caption{MM-CDCR MUC, $B^3$, $CEAFe$ and CoNLL F1 results on AIDA Phase 1 Eval set. Format follows Tables~\ref{tab:mm_cdcr_partial} \&~\ref{tab:mm_cdcr_ensemble}. \LLM~denotes Longformer evaluated with \citet{ahmed20232}'s methodology on novel data. Ensemble model names follow the format {\tt Hard-N model + Hard-P model + Easy pairs model}. \LLM~was always used to handle Easy pairs. The best performing models for hard negative and hard positives were found using a grid search through different combinations of multimodal models. Where only one model is besides \LLM~is listed, that model was used to handle all Hard pairs.}
    \label{tab:mm_cdcr_full_ldc}
\vspace*{-2mm}
\end{table*}